%% file: GaraMoSt-CRC-Arxiv.tex
\documentclass[letterpaper]{article} 
\usepackage{aaai25}  
\usepackage{times}  
\usepackage{helvet}  
\usepackage{courier}  
\usepackage[hyphens]{url}  
\usepackage{graphicx} 
\usepackage{natbib}  
\usepackage{caption} 
\usepackage{algorithm}
\usepackage{algorithmic}

\usepackage[utf8]{inputenc} 
\usepackage[T1]{fontenc}    
\usepackage{url}            
\usepackage{booktabs}       
\usepackage{amsfonts}       
\usepackage{nicefrac}       
\usepackage{microtype}      
\usepackage{xcolor}         

\usepackage{latexsym}
\usepackage{multirow}
\usepackage{bm}
\usepackage{graphicx}
\usepackage{wrapfig2}
\usepackage{natbib}
\setcitestyle{numbers,square}
\usepackage{amssymb}
\usepackage{amsmath}
\usepackage{amsthm}
\usepackage{mathrsfs}
\usepackage{makecell}

\usepackage{colortbl}

\definecolor{GrassGreen}{RGB}{146,208,80}
\definecolor{SkyBlue}{RGB}{4,175,252}
\definecolor{FluoGreen}{RGB}{24,229,25}
\definecolor{Fig1aGreen}{RGB}{106, 234, 102}
\definecolor{Fig1aBlue}{RGB}{0, 176, 240}
\definecolor{EncoderBlue}{RGB}{166, 203, 220}
\definecolor{DecoderGreen}{RGB}{131, 186, 158}
\definecolor{Pending}{RGB}{255, 0, 0}
\definecolor{red}{RGB}{255, 0, 0}
\definecolor{green}{RGB}{0, 255, 0}
\definecolor{blue}{RGB}{0, 0, 255}
\definecolor{AblationRed}{RGB}{251, 221, 238}
\definecolor{AblationRedWord}{RGB}{231, 146, 238}
\definecolor{AblationGreen}{RGB}{193, 220, 206}
\definecolor{AblationGreenWord}{RGB}{160, 191, 118}

\usepackage{newfloat}
\usepackage{listings}
\DeclareCaptionStyle{ruled}{labelfont=normalfont,labelsep=colon,strut=off} 
\floatstyle{ruled}
\newfloat{listing}{tb}{lst}{}
\floatname{listing}{Listing}
\title{GaraMoSt: Parallel Multi-Granularity Motion and Structural Modeling for Efficient Multi-Frame Interpolation in DSA Images}

\author {
    Ziyang Xu\textsuperscript{\rm 1},
    Huangxuan Zhao\textsuperscript{\rm 2,\rm 3,$*$},
    Wenyu Liu\textsuperscript{\rm 1},
    Xinggang Wang\textsuperscript{\rm 1,}\thanks{Corresponding Author.}
}
\affiliations {
    \textsuperscript{\rm 1}Institute of AI, School of EIC, Huazhong University of Science and Technology
	\textsuperscript{\rm 2}School of Computer Science, Wuhan University
    \textsuperscript{\rm 3}Union Hospital, Tongji Medical College, Huazhong University of Science and Technology\\
    \{xuzyoung, liuwy, xgwang\}@hust.edu.cn, zhao\_huangxuan@sina.com
}

\begin{document}

\maketitle

\begin{abstract}
The rapid and accurate direct multi-frame interpolation method for Digital Subtraction Angiography (DSA) images is crucial for reducing radiation and providing real-time assistance to physicians for precise diagnostics and treatment. DSA images contain complex vascular structures and various motions. Applying natural scene Video Frame Interpolation (VFI) methods results in motion artifacts, structural dissipation, and blurriness. Recently, MoSt-DSA has specifically addressed these issues for the first time and achieved SOTA results. However, MoSt-DSA's focus on real-time performance leads to insufficient suppression of high-frequency noise and incomplete filtering of low-frequency noise in the generated images. To address these issues within the same computational time scale, we propose GaraMoSt. Specifically, we optimize the network pipeline with a parallel design and propose a module named MG-MSFE. MG-MSFE extracts frame-relative motion and structural features at various granularities in a fully convolutional parallel manner and supports independent, flexible adjustment of context-aware granularity at different scales, thus enhancing computational efficiency and accuracy. Extensive experiments demonstrate that GaraMoSt achieves the SOTA performance in accuracy, robustness, visual effects, and noise suppression, comprehensively surpassing MoSt-DSA and other natural scene VFI methods. The code and models are available at https://github.com/ZyoungXu/GaraMoSt.
\end{abstract}

\section{Introduction}

4D Digital Subtraction Angiography (DSA) is an advanced medical imaging technology critical for diagnosing and treating various vascular diseases, including in the brain, heart, and limbs \cite{radiation2040028}. It is widely utilized in hospital interventional surgeries. DSA operates by injecting a contrast agent, usually iodine-based, into the patient's body, then capturing images from different angles in space at a fixed rotational speed, alongside the dynamic blood flow changes over time \cite{zhao2024large}. Due to the radiative nature of the imaging environment, the radiation dose received by patients and physicians is directly proportional to the image count, posing a threat to health.

\begin{figure}[t]
    \centering  
    \includegraphics[width=0.45\textwidth]{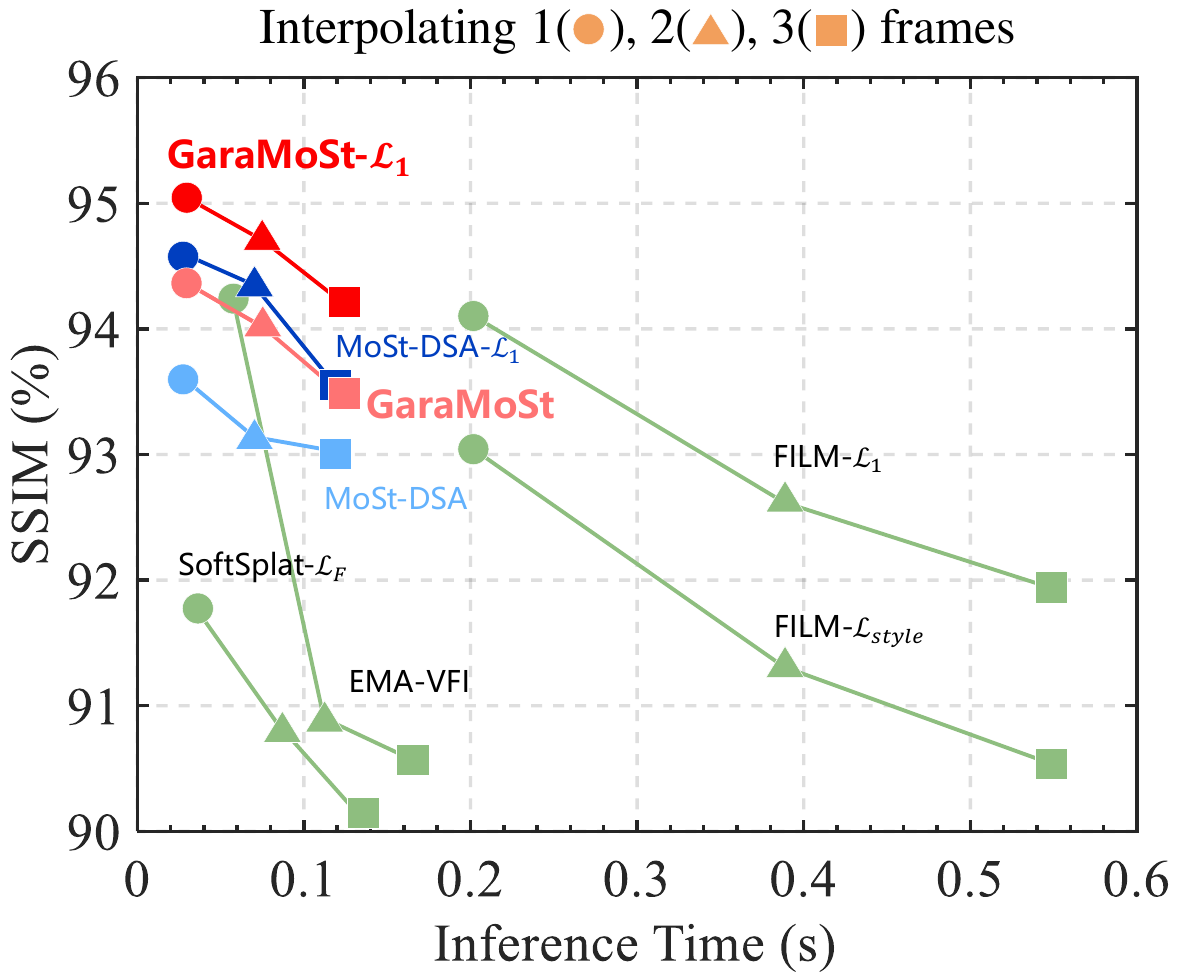}
    \vspace{-2mm}
    \caption{\textbf{SSIM-Time comparison of various methods for interpolating 1 to 3 frames.} Our GaraMoSt-$\mathcal{L}_1$ achieves 95.05, 94.70, 94.22 SSIM, 0.029s, 0.076s, 0.122s inference time, demonstrating SOTA accuracy, while time cost is almost the same as MoSt-DSA. Details in Table \ref{tab:inf1_ave},\ref{tab:inf2_ave},\ref{tab:inf3_ave}.}
    \label{fig:cover}
    \vspace{-7mm}
\end{figure}

\begin{figure*}[t]
    \centering
    \includegraphics[width=0.95\linewidth]{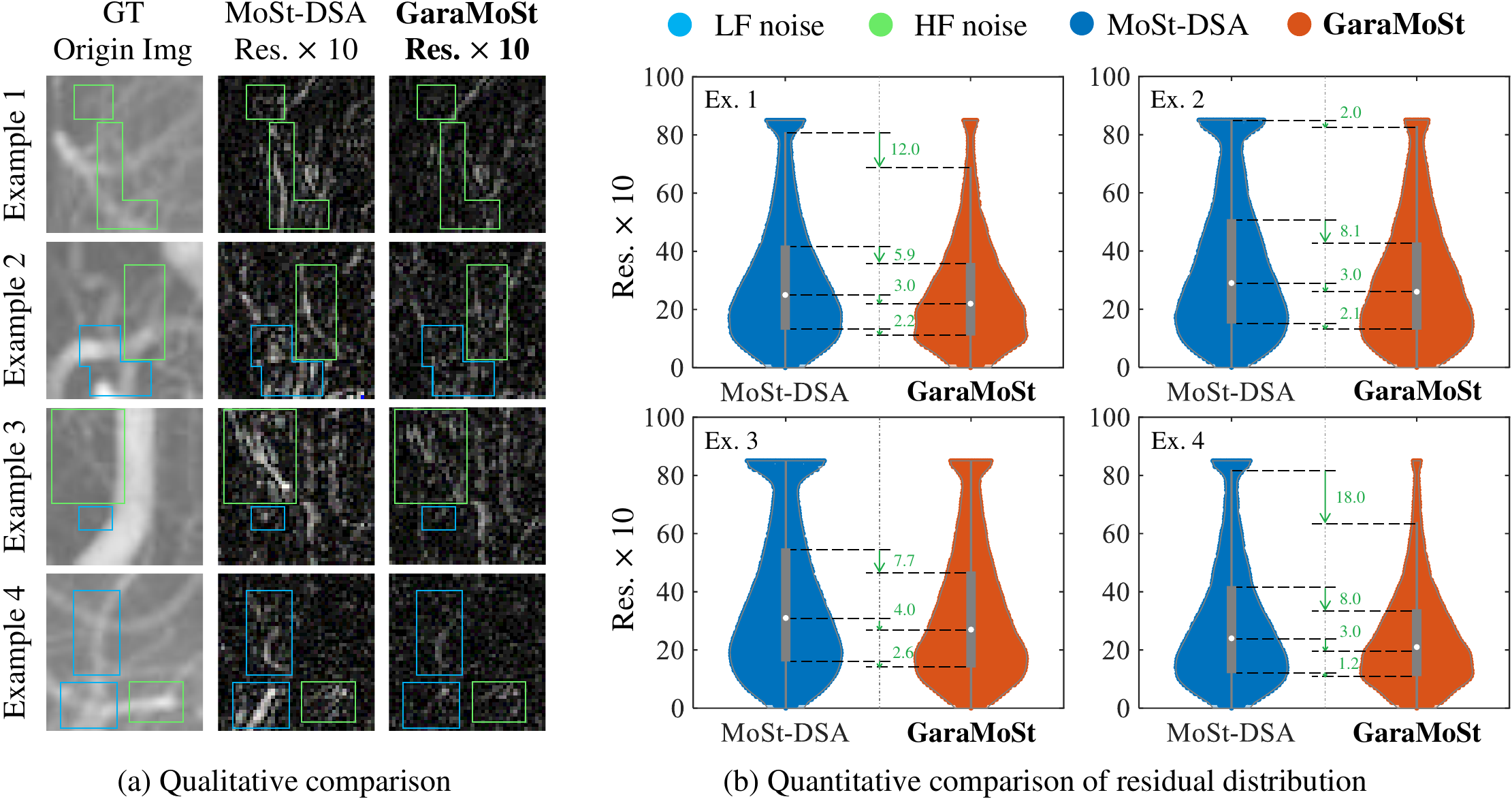}
    \vspace{-2mm}
    \caption{\textbf{Qualitative and quantitative comparison of GaraMoSt vs. MoSt-DSA.} (a) Qualitatively, MoSt-DSA shows insufficient suppression of \textcolor{Fig1aGreen}{high} and \textcolor{Fig1aBlue}{low} frequency noise, whereas GaraMoSt significantly improves these issues. (b) Quantitatively, GaraMoSt enhances noise suppression, with notable reductions in the first quartile, median, third quartile, and upper adjacent of the noise distribution. Additionally, the reduction in width demonstrates a decrease in noise quantity.} 
    \vspace{-4mm}
    \label{fig:intro_noise}
\end{figure*}

Decreasing the number of images through frame interpolation is a straightforward and effective method to reduce radiation. However, simply achieving frame interpolation is insufficient for practical application; the interpolated images must be accurate to assist in diagnostics effectively, and the time cost must be low to aid surgeries in real-time and save valuable patient time. Due to data scarcity and cross-disciplinary complexities, these issues received little attention and lacked targeted solutions. Only recently, MoSt-DSA \cite{Xu2024MoSt-DSA} was the first to address these issues by proposing a deep learning-based direct multi-frame interpolation method. Multi-frame interpolation is a video frame interpolation (VFI) task and a significant research topic in computer vision \cite{Shang_2023_CVPR,jiang2018super,Kalluri_2023_WACV,eva02}. Conventional multi-frame interpolation methods using recursion are time-consuming and unsuitable for real-time support \cite{reda2022film,niklaus2020softmax,park2021abme}. In contrast, MoSt-DSA’s direct multi-frame interpolation method extracts general motion and structural features, maps them at different time steps, and directly infers any number of frames in once forward computation, achieving state-of-the-art (SOTA) performance in terms of image quality, robustness, inference time, and computational cost.

\begin{figure*}[t]
    \centering
    \includegraphics[width=0.9\linewidth]{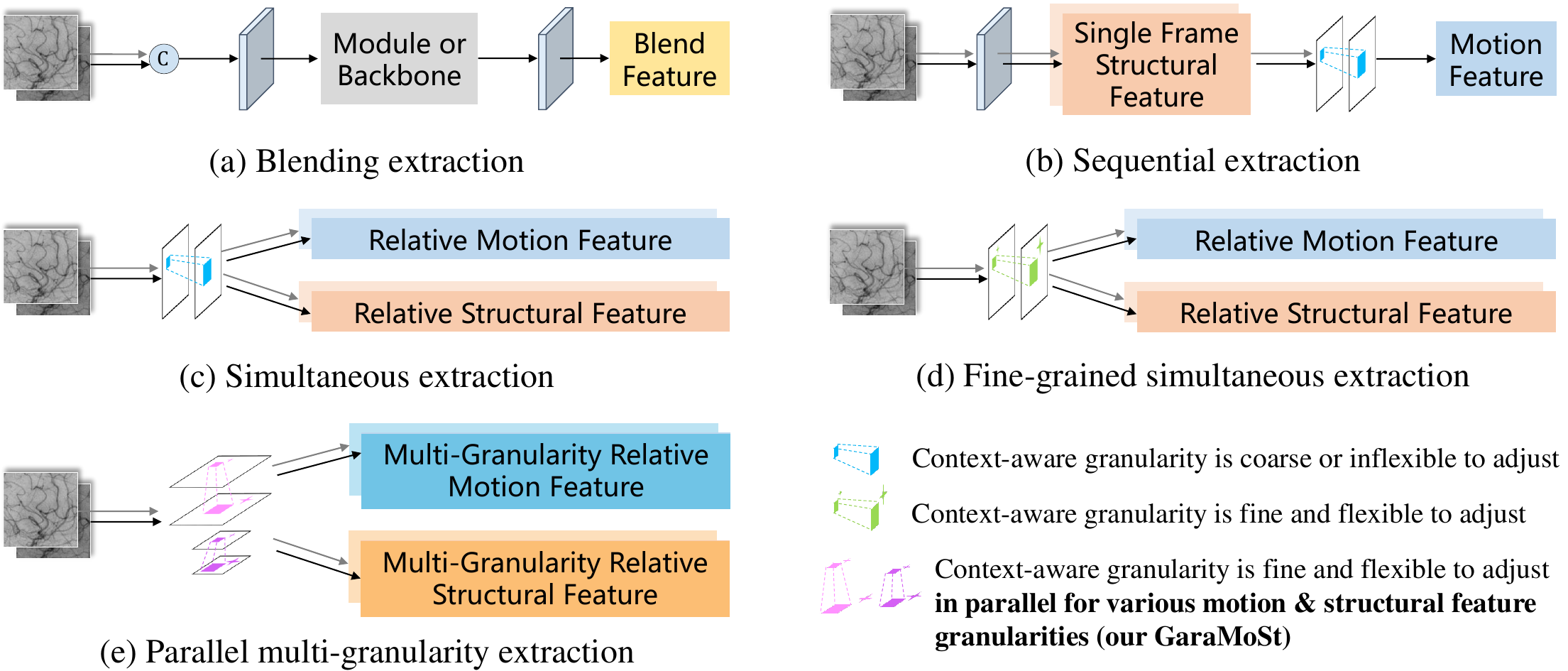}
    \vspace{-2mm}
    \caption{\textbf{Illustration of our proposed parallel multi-granularity extraction and other methods of extracting motion and structural features.} The approach proposed by our GaraMoSt is shown in (e). Notably, "simultaneous" should not be confused with "parallel". While (c) and (d) achieve simultaneous output of motion and structural features outside the module, the internal computation process is still highly sequential.}
    \label{fig:works_type}
    \vskip -0.5 cm
\end{figure*}

The core of the direct multi-frame interpolation method is the precise extraction of motion and structural features through modeling motion and structural interactions \cite{zhang2023extracting}. Existing VFI methods, typically designed for natural scenes, inadequately extract motion and structural features of DSA images or do so at a coarse granularity. Common approaches are categorized into three types, see Figure \ref{fig:works_type} (a), (b), (c): (a) blending motion and structural feature extraction in a single module, leading to ambiguity in both features \cite{Kalluri_2023_WACV,kong2022ifrnet,lu2022video,bao2019depth,ding2021cdfi}; (b) sequentially extracting structural features of each frame and inter-frame motion features in multiple modules, although clearly defining motion features, lack correspondence in inter-frame structure \cite{gao2023video,zhou2023video,yu2023range,danier2022st,jia2022neighbor,niklaus2018context,park2021abme,reda2022film,sim2021xvfi,xue2019video,niklaus2020softmax}; (c) simultaneously extracting relative motion and structural features in a single module, but due to the coarse granularity of context, it fails to adapt to the fine-grained, complex structures of DSA images \cite{zhang2023extracting}. Ultimately, these methods produce motion artifacts, structural dissipation, and blurring issues defined by MoSt-DSA \cite{Xu2024MoSt-DSA}. 
MoSt-DSA introduced a fine-grained context interaction module for simultaneously extracting motion and structural features that enable flexible adjustment of context granularity, significantly improving these issues, as shown in Figure \ref{fig:works_type} (d). However, MoSt-DSA focuses on real-time performance, and although the feature granularity is flexible, the feature sampling scale is limited, leading to insufficient suppression of high-frequency (HF) and low-frequency (LF) noise. As illustrated in Figure \ref{fig:intro_noise}, MoSt-DSA’s inference residuals primarily concentrate on the edges of the tiny blood vessels (HF noise), as well as the sparse background of the vessels and areas with consistent pixel (LF noise).
\vskip -0.1 cm

In summary, a significant challenge is how to further enhance the suppression of HF and LF noise in interpolated images while maintaining low time costs, thereby assisting physicians in more accurate real-time diagnostics and treatment. To address this challenge, we introduce GaraMoSt. Unlike conventional approaches that increase network depth via serial connections for deeper downsampling \cite{zhang2023extracting,huang2020rife}, our method optimizes the network pipeline with a parallel design, using a wider network width to achieve deeper downsampling, maintaining inference time at the same level as MoSt-DSA. Specifically, we first obtain four different scales of features through sequential convolution and downsampling; second, we apply multi-scale dilated convolutions for cross-scale fusion, merging features from layers 1-3 and 2-4 to form two different scales of fused features (1/8 and 1/16); third, we introduce a module named MG-MSFE that extracts relative motion and structural features at various granularities in a fully convolutional parallel manner, enabling flexible adjustment of context-aware granularity across various scales, and transforming optimal context into linear functions, moving away from reliance on expensive attention maps, thereby enhancing computational efficiency and accuracy, as depicted in Figure \ref{fig:works_type} (e); finally, combining different time steps $t$ and the two scales of motion and structural features, we predict dual-layer flows and masks corresponding to the intermediate frame ${I_t}$ and refine it using a simplified UNet \cite{ronneberger2015u} to produce the final intermediate frame ${I_t}$.

In conclusion, our work makes the following contributions:

(1) We introduce GaraMoSt. Compared to MoSt-DSA, GaraMoSt further enhances the suppression of HF and LF noise in interpolated images, with inference time maintained at the same level (for interpolating 3 frames, only increasing by 0.005s), enabling real-time assistance for more precise diagnostic and therapeutic procedures by physicians.

(2) We propose a module named MG-MSFE that extracts frame-relative motion and structural features at various granularities in a fully convolutional parallel manner, supporting independent and flexible adjustment of context-aware granularity across various scales, and transforming optimal context into linear functions, moving away from reliance on expensive attention maps, thereby improving computational efficiency and accuracy.

(3) Extensive experiments demonstrate that GaraMoSt achieves SOTA performance in accuracy, robustness, visual effects, and noise suppression, comprehensively surpassing MoSt-DSA and other natural scene VFI methods.

\begin{figure*}[!t]
    \centering
    \includegraphics[width=0.98\linewidth]{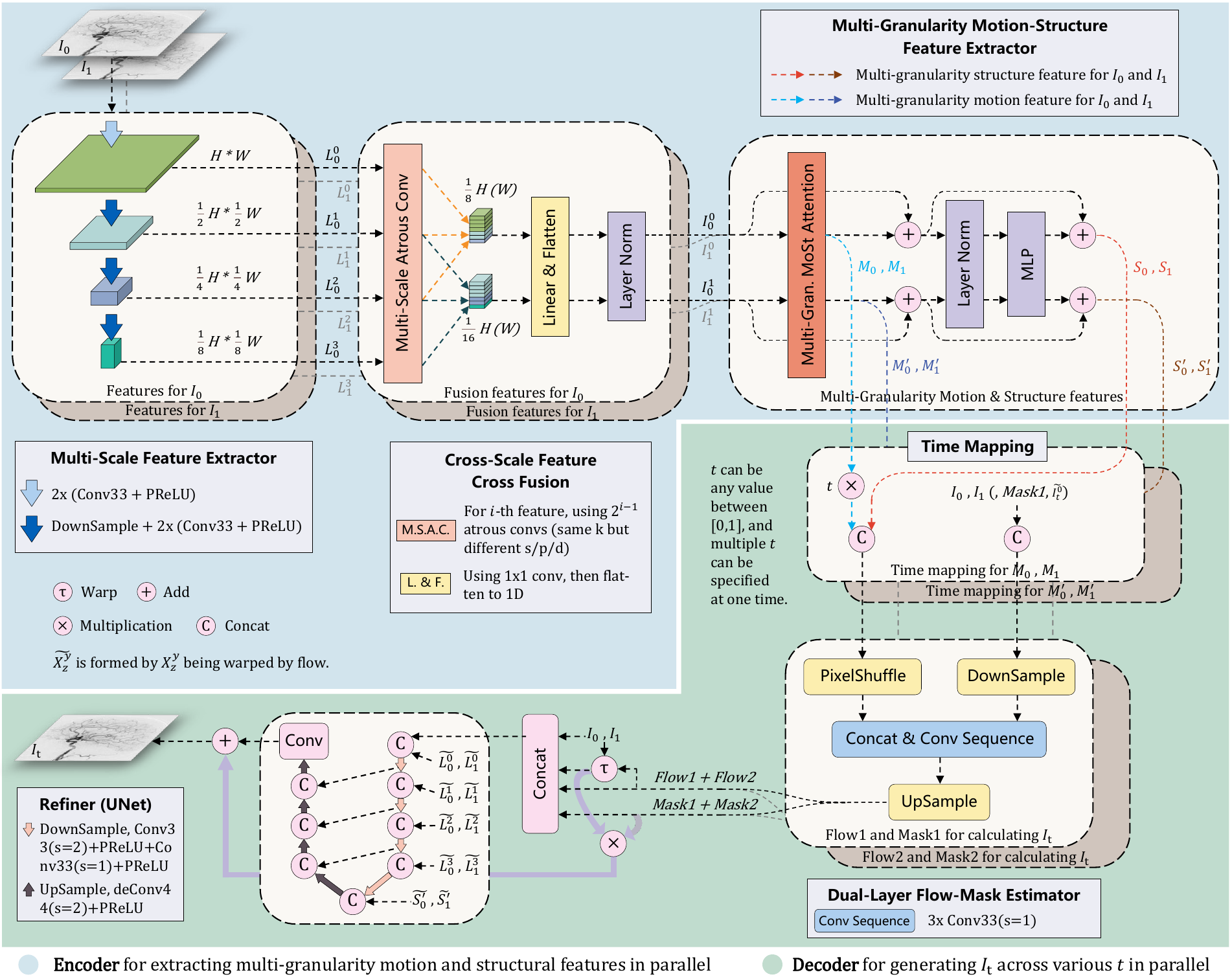}
    \vspace{-2mm}
    \caption{\textbf{Overall pipeline of our GaraMoSt.} The \textbf{\textcolor{EncoderBlue}{encoder}} includes the multi-scale feature extractor (\textbf{MSFE}), cross-scale feature cross fusion (\textbf{CSFCF}), and the multi-granularity motion-structure feature extractor (\textbf{MG-MSFE}) for extracting general multi-granularity motion and structural features in parallel. The \textbf{\textcolor{DecoderGreen}{decoder}} consists of the time mapping (\textbf{TM}), dual-layer flow-mask estimator (\textbf{DL-FME}), and \textbf{Refiner} modules, tailored to parallel decode and refine general multi-granularity features to generate ${I_t}$. Notably, ${t}$ can be any value between $[0,1]$, and multi ${t}$ can be specified at once forward calculation, thus enabling direct multi-frame interpolation in both training and inference.}
    \label{fig:overall_pipeline}
    \vspace{-4mm}
\end{figure*}

\section{Related Work}

\paragraph{Natural scene video frame interpolation.}

SoftSplat \cite{niklaus2020softmax} forward-warp the structural feature pyramid based on an optical flow estimate using softmax splitting. ABME \cite{park2021abme} uses the asymmetric fields to backward warp the input frames' structural features and reconstruct the intermediate frame. FILM \cite{reda2022film} predicts flow at multiple scales to backward warp and fuse structural features. EMA-VFI \cite{zhang2023extracting} extracts relative motion and structural features in a single
module. However, MoSt-DSA's excellent work explains these advanced methods inadequately extract motion and structural features of DSA images or do so at a coarse granularity, producing motion artifacts, structural dissipation, and blurring.

\paragraph{DSA frame interpolation.} 
Characteristics of DSA Images: (1) Filled with microscopic vessels, the fine-grained structure and motion complexity are much higher than in macroscopic natural images; (2) Composed of numerous tiny vessels, whereas natural images are usually composed of countable large instances (e.g., people, cars); (3) Sufficient structural features can usually be obtained with less downsampling; (4) Background pixels are highly similar and uniform. As described in Intro, MoSt-DSA is the first real-time solution but has insufficient suppression of HF and LF noise.

\section{Method}

\begin{figure*}[htbp]
    \centering
    \includegraphics[width=\linewidth]{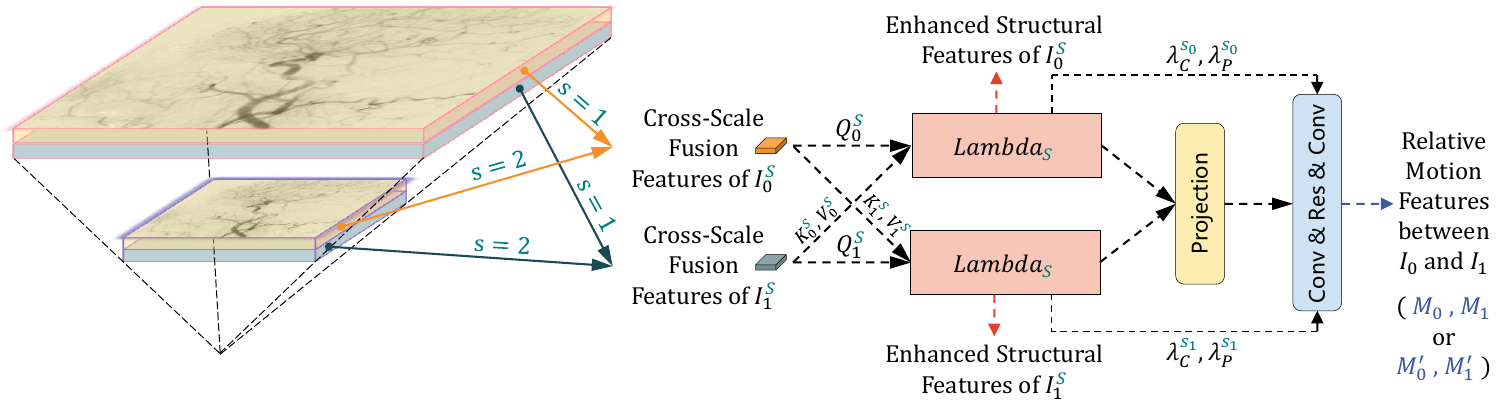}
    \vspace{-7mm}
    \caption{\textbf{Multi-Granularity MoSt-Attention for parallel extracting of multi-granularity motion and structural features.} Different scales correspond to different values of $s$, with $s$ being either $1$ or $2$. In the left image pyramid, the yellow and blue blocks represent the cross-scale fusion features of ${I_0}$ and ${I_1}$ respectively.
    Enhanced structural features are used in MG-MSFE for subsequent calculations to derive the final structural features, see Figure \ref{fig:overall_pipeline} for details.}

    \label{fig:ML-MoSt-Attention}
    \vspace{-3mm}
\end{figure*}

\begin{figure}[t]
    \centering  
    \includegraphics[width=0.28\textwidth]{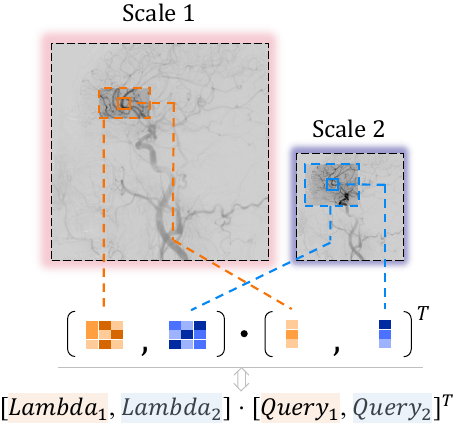}
    \vspace{-2mm}
    \caption{\textbf{Using various granularities in Lambda for feature extraction across scales:} for regions at similar positions across scales, ${Lambda_1}$ and ${Lambda_2}$ summarize different context within different scope $r$ (i.e., granularity) into a fixed-size linear function (i.e., a matrix) applied to the corresponding query, thus bypassing the need for memory-intensive attention maps while maintaining flexibility.}

    \label{fig:Dual_Lambda}
    \vspace{-10mm}
\end{figure}


\paragraph{Improvements Overview.} The pipeline improvements of GaraMoSt over MoSt-DSA include:
\textbf{Feature Extraction}: Added a parallel 1/8 scale feature downsampling for both I0 and I1.
\textbf{Cross-Scale Feature Fusion}: Added a parallel fusion path (Path 1) for 1/2, 1/4, and 1/8 scale features in both I0 and I1. Path 1 runs parallel to Path 2, which fuses 1/1, 1/2, and 1/4 scale features.
\textbf{Motion and Structural Feature Extraction}: Added parallel extraction of relative motion and structural features from Path 1, alongside extraction from Path 2. Moreover, different granularities are adopted for Path 1 and Path 2.
\textbf{Flow and Mask Prediction}: Modified to predict dual-layer flow and mask, with the output as the sum of both layers.
\textbf{Refiner}: Incorporated relative structural features from Path 2 into the lowest-level feature concatenation.

\subsection{Multi-Granularity Motion-Structure Feature Extractor (MG-MSFE)}

As outlined in the introduction, DSA frame interpolation faces challenges in enhancing noise suppression while maintaining low time costs, crucial for accurate real-time diagnostics and treatment. To address this, deepening the feature extraction downsampling is effective but traditional methods \cite{park2021abme,reda2022film,niklaus2020softmax,zhang2023extracting} via serial feature extractor connections are too slow for real-time use. Furthermore, DSA interpolation requires both structural and motion features rich in geometric details. Since deeper neural network layers often lose geometric precision for abstract semantics, using these layers for feature computation and refinement may degrade performance, as our ablation study shows. Hence, we introduce MG-MSFE, parallel extracting multi-granularity relative motion and structural features in shallow layers, enhancing accuracy and reducing noise effectively without sacrificing speed.

\paragraph{Module design and calculation.} 

The complete structure of MG-MSFE is shown in Figure \ref{fig:overall_pipeline}. The input consists of four feature sets: ${\bm{I}_{0}^{0}}$ and ${\bm{I}_{1}^{0}}$ are the cross-scale fusion features of ${\bm{I}_0}$ and ${\bm{I}_1}$ from the 1-3 layers in Multi-Scale Feature Extractor, and ${\bm{I}_{0}^{1}}$ and ${\bm{I}_{1}^{1}}$ are from the 2-4 layers. After encoding these features with Multi-Granularity MoSt Attention, we obtain two different scale motion features, ${\bm{M}_0}$ and ${\bm{M}_{0}^{'}}$ for ${\bm{I}_0}$ relative to ${\bm{I}_1}$, and ${\bm{M}_1}$ and ${\bm{M}_{1}^{'}}$ for ${\bm{I}_1}$ relative to ${\bm{I}_0}$. Simultaneously, Multi-Granularity MoSt Attention encodes the enhanced inter-frame structural features. These initial structural features then pass through residual connections, layer norm, and MLP to obtain the final structural features: ${\bm{S}_0}$, ${\bm{S}_{0}^{'}}$ for ${\bm{I}_0}$ relative to ${\bm{I}_1}$ and ${\bm{S}_1}$, ${\bm{S}_{1}^{'}}$ for ${\bm{I}_1}$ relative to ${\bm{I}_0}$. The sizes of ${\bm{M}_0}$ and ${\bm{M}_1}$ match those of ${\bm{S}_0}$ and ${\bm{S}_1}$, while the sizes of ${\bm{M}_{0}^{'}}$ and ${\bm{M}_{1}^{'}}$ match those of ${\bm{S}_{0}^{'}}$ and ${\bm{S}_{1}^{'}}$.

The calculation process of Multi-Granularity MoSt Attention is shown in Figure \ref{fig:ML-MoSt-Attention}. In the left image pyramid, yellow and blue blocks represent the cross-scale fusion features of ${\bm{I}_0}$ and ${\bm{I}_1}$ respectively. Different scales correspond to $s$ values of $1$ or $2$. First, we employ a Lambda Layer \cite{bello2021lambdanetworks} to simulate content-based and position-based contextual interactions in a fully convolutional manner. Specifically, we denote the depths of the query and value as $|k|$ and $|v|$, respectively, and the position information as $\bm{P}_0^s \in \mathbb{R}^{|n| \times d}$ (each pixel corresponds to a value between [-1.0, 1.0], increasing along rows or columns across different channels). Second, we perform parallel relative attention calculations on the cross-scale fusion features $\bm{I}_0^s \in \mathbb{R}^{|n| \times d}$ and $\bm{I}_1^s \in \mathbb{R}^{|n| \times d}$ for scale $s$. Mathematically, the calculation of $\bm{I}_0^s$ relative to $\bm{I}_1^s$ is as follows, and the calculation of $\bm{I}_1^s$ relative to $\bm{I}_0^s$ is similar:

\begin{align}
\left\{\begin{array}{l}
\bm{Q}_0^s=\bm{I}_0^s \bm{W}_{Q_0^s} \in \mathbb{R}^{|n| \times|k|} \\
\bm{K}_0^s=\bm{I}_1^s \bm{W}_{K_0^s} \in \mathbb{R}^{|n| \times|k|} \\
\bm{V}_0^s=\bm{I}_1^s \bm{W}_{V_0^s} \in \mathbb{R}^{|n| \times|v|}
\end{array}\right..
\end{align}

Then we represent relative position embeddings as $\bm{E}_0^s \in \mathbb{R}^{|n| \times|k|}$ (equivalent to hyperparameters of a 3D-Conv in code implementation). By normalizing the keys, we obtain $\bar{\bm{K}_0^s}=\operatorname{softmax}(\bm{K}_0^s$, axis $=n)$. Next, we compute the content-based contextual interactions $\bm{\lambda}_{c_0}^s$ and position-based contextual interactions $\bm{\lambda}_{p_0}^s$, as:

\vspace{-3mm}
\begin{align}
\left\{\begin{array}{l}
\bm{\lambda}_{c_0}^{s}=\bar{\bm{K}_0^s}^T \bm{V}_0^s \in \mathbb{R}^{|k| \times|v|} \\
\bm{\lambda}_{p_0}^{s}={\bm{E}_0^s}^T \bm{V}_0^s \in \mathbb{R}^{|k| \times|v|}
\end{array}\right..
\end{align}

Finally, by applying contextual interactions to the queries as well as $\bm{P}_0^s$, we obtain the general multi-granularity motion and structural features necessary for inferring any intermediate frame. Specifically, when $s$ takes values of $1$ and $2$, we obtain ${\bm{S}_0}$ and ${\bm{M}_0}$, and ${\bm{S}_0^{'}}$ and ${\bm{M}_0^{'}}$, as:

\vspace{-3mm}
\begin{equation}
\resizebox{0.9\hsize}{!}{$
\left\{\begin{array}{l}
\bm{S}_0=\bm{Q}_0^1 \left( \bm{\lambda}_{c_0}^1+\bm{\lambda}_{p_0}^1\right), {\bm{S}_0^{'}}=\bm{Q}_0^2 \left( \bm{\lambda}_{c_0}^2+\bm{\lambda}_{p_0}^2\right) \\
\bm{M}_0=\bm{P}_0^1 \left( \bm{\lambda}_{c_0}^1+\bm{\lambda}_{p_0}^1\right),
{\bm{M}_0^{'}}=\bm{P}_0^2 \left( \bm{\lambda}_{c_0}^2+\bm{\lambda}_{p_0}^2\right)
\end{array}\right..$}
\end{equation}

All these processes are similar for the calculation of $\bm{I}_1^s$ relative to $\bm{I}_0^s$. In the actual code, we concatenate $\bm{I}_0^s$ and $\bm{I}_1^s$ along the channel dimension to enable parallel computation. Figure \ref{fig:Dual_Lambda} provides a detailed illustration of how to flexibly extract features at different scales with varying granularities. 

\subsection{Parallel Extraction of General Multi-Granularity Motion and Structural Features}

\paragraph{Multi-Scale Feature Extractor (\textbf{MSFE}).}
We use MSFE to obtain vascular features at four different scales. For $\bm{I}_{0}$ and $\bm{I}_{1}$, we first compute the first layer low-level features $\bm{L}_{0}^{0}$ and $\bm{L}_{1}^{0}$ using a 3x3 convolution followed by PReLU \cite{He_2015_ICCV}. Then, through downsampling and the same configuration, we compute the second layer low-level features $\bm{L}_{0}^{1}$ and $\bm{L}_{1}^{1}$. Similarly, we obtain the third and fourth layer low-level features $\bm{L}_{0}^{2}$, $\bm{L}_{1}^{2}$ and $\bm{L}_{0}^{3}$, $\bm{L}_{1}^{3}$. Mathematically,

\vspace{-3mm}
\begin{align}
\left\{\begin{array}{l}
\bm{L}_j^0=\bm{H}\left(\bm{I}_j\right) ,\quad  \bm{L}_j^1=\bm{D}\left(\bm{L}_j^0\right) \\
\bm{L}_j^2=\bm{D}\left(\bm{L}_j^1\right) ,\quad \bm{L}_j^3=\bm{D}\left(\bm{L}_j^2\right) \\
\end{array}\right.,
\end{align}
where $\bm{H}$ is a stack of convolution and activation functions, and $\bm{D}$ represents the integration of $\bm{H}$ and downsampling operations, with $j$ being 0 or 1.

\paragraph{Cross-Scale Feature Cross Fusion (\textbf{CSFCF}).}
We use CSFCF to fuse vascular features. Specifically, we first use multi-scale atrous convolution to fuse the 1-3 layer features and the 2-4 layer features from MSFE. For the $i$-th layer low-level features $\bm{L}_{0}^{i}$ and $\bm{L}_{1}^{i}$, we use $2^{i-1}$ atrous convolutions \cite{chen2017deeplab} (fixed kernel size of 3, stride of $2^{i}$, and for the $n$-th atrous convolution, padding/dilation is $n$). Mathematically,

\vspace{-0.15in}
\begin{align}
\bm{\mathscr{F}}\left(\bm{L}_j^i\right)=\left(\bm{A}_1\left(\bm{L}_j^i\right), \ldots, \bm{A}_n\left(\bm{L}_j^i\right)\right),
\end{align}
where $\bm{\mathscr{F}}$ signifies feature fusion, $\bm{A}$ indicates atrous convolution. The variable $n$, representing the number of $\bm{A}$, takes a value of $2^{i-1}$ ($i$ equal to $0$, $1$, or $2$). By merging the fused features and applying a linear mapping, we obtain four different cross-scale fusion features: ${\bm{I}_{0}^{0}}$ and ${\bm{I}_{1}^{0}}$ from the 1-3 layers, and ${\bm{I}_{0}^{1}}$ and ${\bm{I}_{1}^{1}}$ from the 2-4 layers, as:

\vspace{-0.15in}
\begin{align}
\left\{\begin{array}{l}
\bm{I}_j^{0}=\mathcal{T}\left[\bm{\mathcal{C}}\left(\bm{\mathscr{F}}\left(\bm{L}_j^0\right), \bm{\mathscr{F}}\left(\bm{L}_j^1\right), \bm{\mathscr{F}}\left(\bm{L}_j^2\right)\right)\right] \\
\bm{I}_j^{1}=\mathcal{T}\left[\bm{\mathcal{C}}\left(\bm{\mathscr{F}}\left(\bm{L}_j^1\right), \bm{\mathscr{F}}\left(\bm{L}_j^2\right), \bm{\mathscr{F}}\left(\bm{L}_j^3\right)\right)\right]
\end{array}\right.,
\end{align}
where $j$ being 0 or 1. $\bm{\mathcal{T}}$ represents the linear mapping, and $\bm{\mathcal{C}}$ indicates the concatenation operation. Finally, we flatten and normalize $\bm{I}_j^{0}$ and $\bm{I}_j^{1}$, preparing them for subsequent processing by MG-MSFE.

\subsection{Parallel Decoding and Refinement for Generating Multiple Frames}

\paragraph{Time Mapping (\textbf{TM}).}

We use TM to map the general multi-granularity motion features to the specified time steps in parallel and concatenate them with the corresponding scale general structural features as part of the input ($\alpha$) to the DL-FME. The other part of the input is $\beta$, which varies across different scales. At higher scales, $\beta_{h}$ consists only of the concatenation of $I_0$ and $I_1$, while at lower scales, $\beta_{l}$ includes $I_0$, $I_1$, \scalebox{0.75}{$\widetilde{\bm{I}_t^{0}}$}, and \scalebox{0.75}{$\bm{\mu}_t^{0}$}. Here, $\bm{\mu}_t^{0}$ is the first layer mask calculated by the DL-FME, and \scalebox{0.75}{$\widetilde{\bm{I}_t^{0}}$} is the first layer flow-warped result of ${I_0}$ and ${I_1}$ calculated by the DL-FME.

\paragraph{Dual-Layer Flow-Mask Estimator (\textbf{DL-FME}).}

As shown in Figure \ref{fig:overall_pipeline}, DL-FME (denoted by $\bm{\mathcal{F}}$) applies PixelShuffle \cite{shi2016real} upsampling to $\alpha$, and downsampling to $\beta$. Subsequently, $\alpha$ and $\beta$ merge and undergo continuous convolution operations, eventually generating bidirectional optical flow $\bm{\phi}_t^{s}$ and mask $\bm{\mu}_t^{s}$ for the specific $t$ through upsampling in different layers, as:

\vspace{-0.15in}
\begin{align}
\left\{\begin{array}{l}
\bm{\phi}_t^{0}, \bm{\mu}_t^{0}=\bm{\mathcal{F}} \left(\bm{\mathcal{C}}\left(\bm{M}_{0 \rightarrow t}^{0}, \bm{M}_{1 \rightarrow t}^{0}, \bm{S}_0^{0}, \bm{S}_1^{0}\right), \beta_{h}\right) \\
\bm{\phi}_t^{1}, \bm{\mu}_t^{1}=\bm{\mathcal{F}} \left(\bm{\mathcal{C}}\left(\bm{M}_{0 \rightarrow t}^{1}, \bm{M}_{1 \rightarrow t}^{1}, \bm{S}_0^{1}, \bm{S}_1^{1}\right), \beta_{l}\right)
\end{array}\right..
\end{align}

Next, we sum $\bm{\phi}_t$ and $\bm{\mu}_t$ across different layers, using the summed $\bm{\phi}_t$ to warp $\bm{I}_0$, $\bm{I}_1$, the low-level features $\bm{L}_j^i$, and \scalebox{0.75}{$\bm{S}_0^{'}$} and \scalebox{0.75}{$\bm{S}_1^{'}$}. For instance, for $\bm{X}_z^y$, the result after warping is denoted as \scalebox{0.75}{$\widetilde{\bm{X}_z^y}$}. Subsequently, we concatenate $\bm{I}_0$, $\bm{I}_1$, \scalebox{0.75}{$\widetilde{\bm{I}_0}$}, \scalebox{0.75}{$\widetilde{\bm{I}_1}$}, $\bm{\phi}_t$, and $\bm{\mu}_t$ together, referred to as $\mathcal{O}_t$ for Refiner to process.

\paragraph{\textbf{Refiner}.}
Follow the design of MoSt-DSA, except incorporated \scalebox{0.75}{$\widetilde{\bm{S}_0^{'}}$} and \scalebox{0.75}{$\widetilde{\bm{S}_1^{'}}$} into the lowest-level feature concatenation.

\section{Experiments}

\paragraph{Model Configuration.}
For the best accuracy-time trade-off, context-aware granularities ($r$) are $[7,7]$, $[7,29]$, $[7,29]$. See the ablation study for proof.

\paragraph{Training Details.}
We followed the settings of MoSt-DSA \cite{Xu2024MoSt-DSA}, except for the following differences. For interpolating 1 to 3 frames, we set the batch sizes to 10, 6, and 4, with 1000 warm-up steps. We use the AdamW \cite{loshchilov2018fixing} optimizer with $\beta_1=0.9$, $\beta_2=0.999$, and a weight decay of $6e-5$. The learning rate is warmed up to $6e-5$ and then decays to $6e-6$ over 100 epochs following a cosine schedule \cite{loshchilov2016sgdr}.

\paragraph{Comparison Details.}
For a fair comparison, we followed MoSt-DSA's settings, including trained two versions on their DSA dataset: one (GaraMoSt-$\mathcal{L}_1$) using only the $\mathcal{L}_1$ loss, which achieves higher test scores, and the other (GaraMoSt) using the combined loss $\mathcal{L}$ \cite{Xu2024MoSt-DSA}, which benefits image quality. MoSt-DSA's dataset \cite{Xu2024MoSt-DSA} contains 470 head DSA image sequences (329 for training and 141 for testing; 489*489 resolution) from eight hospitals, which are made in the data form of interpolating 1 to 3 frames. Our evaluation results are fairly derived from a single NVIDIA RTX 3090 GPU.

\input{tables/Inf1_ave}
\input{tables/Inf2_ave.tex}

\input{tables/Inf3_ave.tex}

\begin{figure*}[t]
    \centering
    \includegraphics[width=\linewidth]{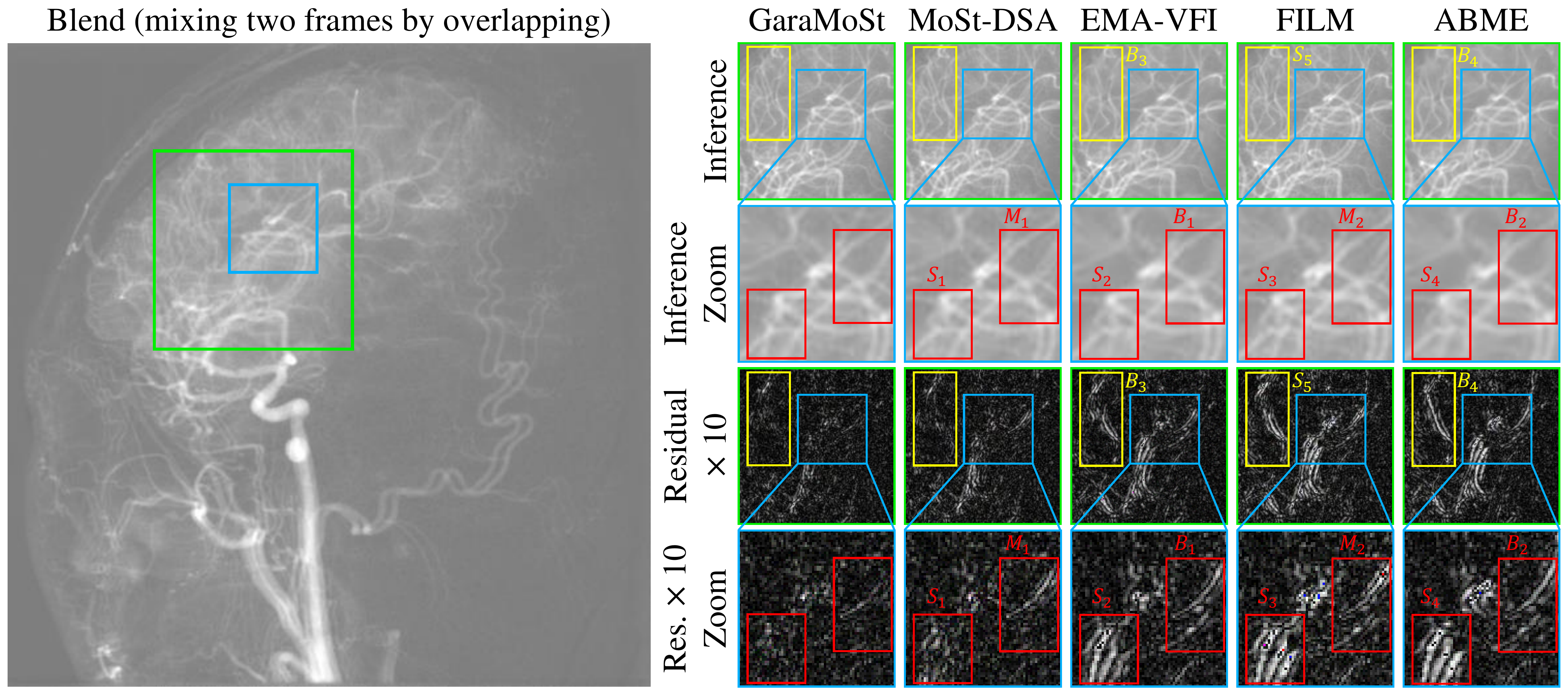}
    \vspace{-6mm}
    \caption{\textbf{Visual comparison for interpolating one frame: our GaraMoSt vs. other SOTA methods.} The first and third rows correspond to the \textbf{\textcolor{green}{green}} box in the "blend", while the second and fourth rows correspond to the \textbf{\textcolor{SkyBlue}{blue}} box in the "blend". $M$ stands for motion artifact, $S$ for structural dissipation, and $B$ for blurring. The residual levels of our GaraMoSt are significantly lower than those of all other methods. This demonstrates that we effectively suppressed noise and further improved issues related to motion artifacts, structural dissipation, and blurring.}
    \label{fig:visual_compare}
    \vspace{-4mm}
\end{figure*}

\subsection{Single-Frame Interpolation}

The quantitative evaluation results are shown in Table 1. EMA is short for EMA-VFI \cite{zhang2023extracting}. Our method achieved significant accuracy improvements whether compared with methods using color losses or perceptually-sensitive losses. Moreover, the substantial reduction in the STD of SSIM (S.SSIM) demonstrates our model's significant robustness enhancement. Despite leading in accuracy, our inference time is nearly identical to MoSt-DSA, with only a 0.005s difference. Specifically, compared to methods using color losses, GaraMoSt-$\mathcal{L}_1$ achieved a Mean of SSIM (M.SSIM) of 95.05 (\textbf{+0.33}), an S.SSIM of 1.89 (\textbf{-0.23}), a Mean of PSNR (M.PSNR) of 40.69 (\textbf{+0.37}), and a STD of PSNR (S.PSNR) of 3.29 (\textbf{-0.06}). Compared to methods using perceptually-sensitive losses, GaraMoSt performed even better, achieving an M.SSIM of 94.37 (\textbf{+0.72}), an S.SSIM of 2.22 (\textbf{-0.39}), an M.PSNR of 40.19 (\textbf{+0.64}), and an S.PSNR of 3.31 (\textbf{-0.14}).
Notably, the score differences in SSIM and PSNR among SOTA VFI methods are minimal. For instance, on the UCF101 \cite{soomro2012ucf101}, the SOTA EMA-VFI surpasses the second-best \cite{jin2023unified} by only 0.01\% in SSIM and 0.01 in PSNR, and the third-best \cite{zhou2023video} by 0.04\% in SSIM and 0.01 in PSNR. Thus, the superiority of our method is significant.

The qualitative comparison results are shown in Figure \ref{fig:visual_compare}.
The residual levels of our GaraMoSt are significantly lower than those of all other methods. This demonstrates that we effectively suppressed noise and further improved issues related to motion blur, structural dissipation, and blurring.

\subsection{Multi-Frame Interpolation}

The quantitative results for interpolating 2 and 3 frames are presented in Table \ref{tab:inf2_ave}, \ref{tab:inf3_ave}. We continue to show significant improvements in accuracy and robustness. The inference time for interpolating 2 and 3 frames remains almost the same as MoSt-DSA, with differences of 0.006s and 0.005s. Notably, when interpolating 2 frames, our GaraMoSt achieves an M.SSIM of 94.01 (\textbf{+0.87}), an S.SSIM of 2.42 (\textbf{-0.42}), and an M.PSNR of 39.66 (\textbf{+0.72}). When interpolating 3 frames, our GaraMoSt-$\mathcal{L}_1$ achieves an M.SSIM of 94.22 (\textbf{+0.64}), an S.SSIM of 2.34 (\textbf{-0.35}), and an M.PSNR of 39.53 (\textbf{+0.68}).

\subsection{Ablation Study}

\paragraph{Impact of various context-aware granularity in Single-Frame Interpolation.} 

As shown in Supplementary Material's Table \ref{tab:ablation_inf1}, for the best accuracy-time trade-off, $[7,7]$, $[15,15]$, $[7,29]$, and $[15,29]$ are the best 4 options, while $[29,15]$ is the worst.
Among various schemes, the time differences up to $0.016$s, M.SSIM differences up to $0.19$, and S.SSIM differences up to $0.07$. This demonstrates the importance of choosing the appropriate granularity.

\paragraph{Impact of various context-aware granularity in Multi-Frame Interpolation.} We further compared the single-frame interpolation's best 4 options on multi-frame interpolation (MFI), as shown in Supplementary Material's Table \ref{tab:SP-Ablation-multi}. The results indicate that $[7,29]$ is more suitable for MFI.

\paragraph{Influence of structural features with different depths on refinement.} 

We upsampled \scalebox{0.75}{$\widetilde{\bm{S}_0^{'}}$} and \scalebox{0.75}{$\widetilde{\bm{S}_1^{'}}$} to match the size of \scalebox{0.75}{$\widetilde{\bm{L}_0^{3}}$} and \scalebox{0.75}{$\widetilde{\bm{L}_1^{3}}$} and replaced \scalebox{0.75}{$\widetilde{\bm{L}_0^{3}}$} and \scalebox{0.75}{$\widetilde{\bm{L}_1^{3}}$} in the Refiner with the upsampled features, resulting in a decrease of $0.10$ in M.SSIM, and a decrease of $0.09$ in M.PSNR. This indicates the importance of using shallow structural features for refinement. 
See details in Supplementary Material's Table \ref{tab:SP-ablation-depth}.

\section{Conclusion}

We proposed GaraMoSt, which further enhances noise suppression in interpolated images while maintaining low time costs, enabling real-time assistance for more precise diagnostic and therapeutic procedures by physicians.
In particular, we devised a general module MG-MSFE that efficiently extracts relative motion and structural features at various granularities in a fully convolutional parallel manner, supporting independent and flexible adjustment of context-aware granularity across various scales. Extensive experiments show that our GaraMoSt achieves SOTA performance in accuracy, robustness, visual effects, and noise suppression, comprehensively surpassing MoSt-DSA and other natural scene VFI methods.


\newpage
\section{Acknowledgments}
This work is supported by the National Natural Science Foundation of China (No. 62376102, No. 82472070).

\bibliography{aaai25-CRC-Arxiv}

\clearpage
\section{Supplementary Materials}
\subsection{Impact of various context-aware granularity in Single-Frame Interpolation} 

As shown in Table \ref{tab:ablation_inf1}, for the best accuracy-time trade-off, $[7,7]$, $[15,15]$, $[7,29]$, and $[15,29]$ are the best 4 options, while $[29,15]$ is the worst. 
Among various schemes, the time differences up to $0.016$s, M.SSIM differences up to $0.19$, and S.SSIM differences up to $0.07$. This demonstrates the importance of choosing the appropriate granularity.

\input{tables/ablation_inf1.tex}

\subsection{Impact of various context-aware granularity in Multi-Frame Interpolation} We further compared the single-frame interpolation's best 4 options on multi-frame interpolation, as shown in Table \ref{tab:SP-Ablation-multi}. The results indicate that $[7,29]$ is more suitable for multi-frame interpolation.

\input{tables/SP-ablation-multi.tex}

\subsection{Influence of structural features with different depths on refinement}

We upsampled \scalebox{0.75}{$\widetilde{\bm{S}_0^{'}}$} and \scalebox{0.75}{$\widetilde{\bm{S}_1^{'}}$} to match the size of \scalebox{0.75}{$\widetilde{\bm{L}_0^{3}}$} and \scalebox{0.75}{$\widetilde{\bm{L}_1^{3}}$} and replaced \scalebox{0.75}{$\widetilde{\bm{L}_0^{3}}$} and \scalebox{0.75}{$\widetilde{\bm{L}_1^{3}}$} in the Refiner with the upsampled features, for single-frame interpolation. The result is shown in Table \ref{tab:SP-ablation-depth}, M.SSIM decreased by $0.10$, M.PSNR decreased by $0.09$, and the increase in S.SSIM and S.PSNR indicates a reduction in robustness. This demonstrates the importance of using shallow structural features for refinement, as using deeper structural features with higher semantic information density may not be suitable for refinement.

\input{tables/SP-ablation-depth.tex}

\end{document}

%% file: tables/Inf1_ave.tex
\begin{table}[t]
\centering
\caption{\textbf{Quantitative comparison with SOTA methods on \textbf{single-frame} interpolation.} Best scores for color losses in \textbf{\textcolor{blue}{blue}}, and for perceptually-sensitive losses in \textbf{\textcolor{red}{red}}. “$\dagger$” obtained by ourselves, the rest are copied from \cite{Xu2024MoSt-DSA}.}
\vspace{-2mm}
\resizebox{0.45\textwidth}{!}{
    \begin{tabular}{cccccc}
    \toprule
    \multirow{2}[3]{*}{\textbf{Method}} & \multicolumn{2}{c}{\textbf{SSIM ($\%$)}} & \multicolumn{2}{c}{\textbf{PSNR}} & \textbf{Time (s)} \\
    \cmidrule{2-6}       & Mean$\uparrow$ & STD$\downarrow$  & Mean$\uparrow$ & STD$\downarrow$  & 1 frame$\downarrow$ \\
    \midrule
    ABME \cite{park2021abme} & 94.02  & 2.28 & 39.83 & 3.39  & 0.383   \\
    FILM-$\mathcal{L}_1$ \cite{reda2022film} & 94.11  & 2.37 & 39.86 & 3.43  & 0.201   \\
    EMA-small \cite{zhang2023extracting} & 94.19  & 2.21 & 40.07 & 3.47  & 0.027   \\
    EMA \cite{zhang2023extracting}   & 94.33  & 2.13 & 40.13 & 3.40  & 0.056    \\
    MoSt-DSA-$\mathcal{L}_1$ \cite{Xu2024MoSt-DSA} & 94.62  & 2.12 & 40.32 & 3.35  & 0.024    \\
    GaraMoSt-$\mathcal{L}_1$$\dagger$ & \textbf{\textcolor{blue}{95.05}}  & \textbf{\textcolor{blue}{1.89}} & \textbf{\textcolor{blue}{40.69}} & \textbf{\textcolor{blue}{3.29}}  & 0.029    \\
    \midrule
    SoftSplat-$\mathcal{L}_F$ \cite{niklaus2020softmax} & 91.78  & 3.07 & 38.59 & 3.51  & 0.035   \\
    FILM-$\mathcal{L}_{VGG}$ \cite{reda2022film} & 93.10  & 2.67 & 39.27 & 3.45  & 0.201  \\
    FILM-$\mathcal{L}_{Style}$ \cite{reda2022film} & 93.05  & 2.72 & 39.25 & 3.48  & 0.201   \\
    MoSt-DSA \cite{Xu2024MoSt-DSA} & 93.65  & 2.61 & 39.55 & 3.45  & 0.024   \\
    GaraMoSt$\dagger$ & \textbf{\textcolor{red}{94.37}}  & \textbf{\textcolor{red}{2.22}} & \textbf{\textcolor{red}{40.19}} & \textbf{\textcolor{red}{3.31}}  & 0.029   \\
    \bottomrule
    \end{tabular}%
}
\label{tab:inf1_ave}%
\end{table}%

%% file: tables/Inf2_ave.tex
\begin{table}[t]
\centering
\caption{\textbf{Quantitative comparison with SOTA methods on \textbf{two frames} interpolation.} \textbf{\textcolor{blue}{Blue}}, \textbf{\textcolor{red}{red}}, $\dagger$ same to Table \ref{tab:inf1_ave}.}
\vspace{-2mm}
\resizebox{0.45\textwidth}{!}{
    \begin{tabular}{cccccc}
    \toprule
    \multirow{2}[3]{*}{\textbf{Method}} & \multicolumn{2}{c}{\textbf{SSIM ($\%$)}} & \multicolumn{2}{c}{\textbf{PSNR}} & \textbf{Time (s)} \\
    \cmidrule{2-6}       & Mean$\uparrow$ & STD$\downarrow$  & Mean$\uparrow$ & STD$\downarrow$  & 2 frame$\downarrow$ \\
    \midrule
    FILM-$\mathcal{L}_1$ \cite{reda2022film} & 92.62 & 3.13 & 37.93 & 3.82  & 0.388  \\
    EMA-small \cite{zhang2023extracting} & 91.82 & 3.68 & 37.37 & 4.07  & 0.074  \\
    EMA \cite{zhang2023extracting} & 91.90 & 3.63 & 37.41 & 4.08  & 0.112   \\
    MoSt-DSA-$\mathcal{L}_1$ \cite{Xu2024MoSt-DSA} & 94.35 & 2.29 & 39.78 & 3.44 & 0.070  \\
    GaraMoSt-$\mathcal{L}_1$$\dagger$ & \textbf{\textcolor{blue}{94.70}} & \textbf{\textcolor{blue}{2.08}} & \textbf{\textcolor{blue}{40.16}} & \textbf{\textcolor{blue}{3.36}} & 0.076  \\
    \midrule
    SoftSplat-$\mathcal{L}_F$ \cite{niklaus2020softmax} & 90.86 & 3.49 & 37.84 & 3.43  & 0.084   \\
    FILM-$\mathcal{L}_{VGG}$ \cite{reda2022film} & 91.42 & 3.42 & 37.47 & 3.67  & 0.388   \\
    FILM-$\mathcal{L}_{Style}$ \cite{reda2022film} & 91.31 & 3.50 & 37.39 & 3.74  & 0.388  \\
    MoSt-DSA \cite{Xu2024MoSt-DSA} & 93.14 & 2.84 & 38.94 & 3.38 & 0.070  \\
    GaraMoSt$\dagger$ & \textbf{\textcolor{red}{94.01}} & \textbf{\textcolor{red}{2.42}} & \textbf{\textcolor{red}{39.66}} & \textbf{\textcolor{red}{3.37}} & 0.076  \\
    \bottomrule
    \end{tabular}%
}
\label{tab:inf2_ave}%
\vspace{-4mm}
\end{table}%

%% file: tables/Inf3_ave.tex
\vspace{-3mm}
\begin{table}[H]
\centering
\caption{\textbf{Quantitative comparison with SOTA methods on \textbf{three frames} interpolation.} \textbf{\textcolor{blue}{Blue}}, \textbf{\textcolor{red}{red}}, $\dagger$ same to Table \ref{tab:inf1_ave}.}

\vspace{-2mm}
\resizebox{0.45\textwidth}{!}{
    \begin{tabular}{cccccc}
    \toprule
    \multirow{2}[3]{*}{\textbf{Method}} & \multicolumn{2}{c}{\textbf{SSIM ($\%$)}} & \multicolumn{2}{c}{\textbf{PSNR}} & \textbf{Time (s)} \\
    \cmidrule{2-6}       & Mean$\uparrow$ & STD$\downarrow$  & Mean$\uparrow$ & STD$\downarrow$  & 3 frame$\downarrow$ \\
    \midrule
    FILM-$\mathcal{L}_1$ \cite{reda2022film} & 91.94 & 3.52 & 37.21 & 3.91  & 0.548  \\
    EMA-small \cite{zhang2023extracting} & 90.48 & 4.52 & 36.31 & 4.24  & 0.122   \\
    EMA \cite{zhang2023extracting} & 90.57 & 4.50 & 36.35 & 4.24  & 0.165  \\
    MoSt-DSA-$\mathcal{L}_1$ \cite{Xu2024MoSt-DSA} & 93.58 & 2.69 & 38.85 & 3.56 & 0.117 \\
    GaraMoSt-$\mathcal{L}_1$$\dagger$ & \textbf{\textcolor{blue}{94.22}} & \textbf{\textcolor{blue}{2.34}} & \textbf{\textcolor{blue}{39.53}} & \textbf{\textcolor{blue}{3.51}} & 0.122  \\
    \midrule
    SoftSplat-$\mathcal{L}_F$ \cite{niklaus2020softmax} & 90.07 & 3.87 & 37.24 & 3.53 & 0.137  \\
    FILM-$\mathcal{L}_{VGG}$ \cite{reda2022film} & 90.63 & 3.83 & 36.75 & 3.74  & 0.548 \\
    FILM-$\mathcal{L}_{Style}$ \cite{reda2022film} & 90.54 & 3.90 & 36.66 & 3.82  & 0.548   \\
    MoSt-DSA \cite{Xu2024MoSt-DSA} & 93.03 & 2.94 & 38.66 & 3.59 & 0.117   \\
    GaraMoSt$\dagger$ & \textbf{\textcolor{red}{93.49}} & \textbf{\textcolor{red}{2.71}} & \textbf{\textcolor{red}{39.06}} & \textbf{\textcolor{red}{3.48}} & 0.122  \\
    \bottomrule
    \end{tabular}%
}
\label{tab:inf3_ave}%
\vspace{-2mm}
\end{table}%

%% file: tables/ablation_inf1.tex

\begin{table}[htbp]
\centering
\caption{\textbf{Ablation of various context-aware granularity on single-frame interpolation.} The best 4 strategies are in \textbf{\textcolor{AblationGreenWord}{green}}, and the worst is in \textbf{\textcolor{AblationRedWord}{pink}}.}
\resizebox{0.45\textwidth}{!}{
    \begin{tabular}{cccccc}
    \toprule
    \multicolumn{1}{c}{\textbf{Granularity}} & \multicolumn{2}{c}{\textbf{SSIM (\%)}} & \multicolumn{2}{c}{\textbf{PSNR}} & \textbf{Time (s)} \\
    \cmidrule{2-6}    [$level1, level2$] & Mean↑ & STD↓  & Mean↑ & STD↓  & 1 frame↓ \\
    \midrule
    \rowcolor{AblationGreen} 7, 7  & 94.37  & 2.22  & 40.19  & 3.31  & 0.029  \\
    \rowcolor{AblationGreen} 15, 15 & 94.31  & 2.25  & 40.11  & 3.31  & 0.030  \\
    21, 21 & 94.28  & 2.25  & 40.10  & 3.28  & 0.036  \\
    29, 29 & 94.28  & 2.28  & 40.11  & 3.33  & 0.045  \\
    \midrule
    7, 15 & 94.24  & 2.29  & 40.08  & 3.34  & 0.030  \\
    7, 21 & 94.30  & 2.26  & 40.13  & 3.33  & 0.030  \\
    \rowcolor{AblationGreen} 7, 29 & 94.34  & 2.24  & 40.16  & 3.31  & 0.031  \\
    15, 21 & 94.24  & 2.29  & 40.08  & 3.33  & 0.034  \\
    \rowcolor{AblationGreen} 15, 29 & 94.33  & 2.25  & 40.15  & 3.31  & 0.035  \\
    21, 29 & 94.25  & 2.26  & 40.09  & 3.31  & 0.038  \\
    \midrule
    15, 7 & 94.26  & 2.25  & 40.09  & 3.30  & 0.032  \\
     21, 7 & 94.28  & 2.27  & 40.12  & 3.33  & 0.034  \\
    21, 15 & 94.23  & 2.29  & 40.07  & 3.33  & 0.035  \\
    29, 7 & 94.28  & 2.25  & 40.10  & 3.27  & 0.042  \\
    \rowcolor{AblationRed} 29, 15 & 94.18  & 2.29  & 40.07  & 3.34  & 0.043  \\
    29, 21 & 94.26  & 2.25  & 40.11  & 3.30  & 0.043  \\
    \bottomrule
    \end{tabular}%
}
\label{tab:ablation_inf1}%
\end{table}%

%% file: tables/SP-ablation-multi.tex
\begin{table}[htbp]
  \centering
  \caption{\textbf{Comparison of the performance of the 4 best single-frame interpolation options in multi-frame interpolation.} The best strategies is in \textbf{\textcolor{AblationGreenWord}{green}}.}
  \resizebox{0.47\textwidth}{!}{
    \begin{tabular}{ccccccc}
    \toprule
    \multirow{2}[4]{*}{} & \multicolumn{1}{c}{\textbf{Granularity}} & \multicolumn{2}{c}{\textbf{SSIM (\%)}} & \multicolumn{2}{c}{\textbf{PSNR}} & \textbf{Time (s)} \\
    \cmidrule{3-7}          & [$level1, level2$] & Mean↑ & STD↓  & Mean↑ & STD↓  & $n$ frame↓ \\
    \midrule
    \multirow{4}[2]{*}{\makecell[c]{Single-\\Frame\\Interp.}} & \cellcolor{AblationGreen} 7, 7  & \cellcolor{AblationGreen}94.37  & \cellcolor{AblationGreen}2.22  & \cellcolor{AblationGreen}40.19  & \cellcolor{AblationGreen}3.31  & \cellcolor{AblationGreen}0.029  \\
          & 15, 15 & 94.31  & 2.25  & 40.11  & 3.31  & 0.030  \\
          & 7, 29 & 94.34  & 2.24  & 40.16  & 3.31  & 0.031  \\
          & 15, 29 & 94.33  & 2.25  & 40.15  & 3.31  & 0.035  \\
    \midrule
    \multirow{4}[2]{*}{\makecell[c]{Two-\\Frame\\Interp.}} & 7, 7  & 93.94  & 2.45  & 39.62  & 3.37  & 0.075  \\
          & 15, 15 & 93.97  & 2.42  & 39.63  & 3.38  & 0.078  \\
          & \cellcolor{AblationGreen} 7, 29 & \cellcolor{AblationGreen}94.01  & \cellcolor{AblationGreen}2.42  & \cellcolor{AblationGreen}39.66  & \cellcolor{AblationGreen}3.37  & \cellcolor{AblationGreen}0.076  \\
          & 15, 29 & 93.99  & 2.42  & 39.66  & 3.38  & 0.080  \\
    \midrule
    \multirow{4}[2]{*}{\makecell[c]{Three-\\Frame\\Interp.}} & 7, 7  & 93.41  & 2.74  & 39.03  & 3.49  & 0.121  \\
          & 15, 15 & 93.42  & 2.74  & 39.01  & 3.49  & 0.125  \\
          & \cellcolor{AblationGreen} 7, 29 & \cellcolor{AblationGreen}93.49  & \cellcolor{AblationGreen}2.71  & \cellcolor{AblationGreen}39.06  & \cellcolor{AblationGreen}3.48  & \cellcolor{AblationGreen}0.122  \\
          & 15, 29 & 93.45  & 2.72  & 39.04  & 3.48  & 0.127  \\
    \bottomrule
    \end{tabular}
    }
  \label{tab:SP-Ablation-multi}
\end{table}%

%% file: tables/SP-ablation-depth.tex
\begin{table}[htbp]
  \centering
  \caption{\textbf{Comparison of using structural features with different depths in Refiner.}}
  \resizebox{0.47\textwidth}{!}{
    \begin{tabular}{ccccccc}
    \toprule
    \multicolumn{1}{c}{\textbf{Granularity}} & \multicolumn{1}{c}{\multirow{2}[4]{*}{\textbf{Sturcture Feature}}} & \multicolumn{2}{c}{\textbf{SSIM (\%)}} & \multicolumn{2}{c}{\textbf{PSNR}} & \textbf{Time (s)} \\
\cmidrule{3-7}    [$level1, level2$] &       & Mean↑ & STD↓  & Mean↑ & STD↓  & 1 frame↓ \\
    \midrule
    7, 7  & \scalebox{0.75}{$\widetilde{\bm{L}_0^{3}}$}, \scalebox{0.75}{$\widetilde{\bm{L}_1^{3}}$} & 94.37  & 2.22  & 40.19  & 3.31  & 0.029  \\
    7, 7  & upsampled \scalebox{0.75}{$\widetilde{\bm{S}_0^{'}}$}, \scalebox{0.75}{$\widetilde{\bm{S}_1^{'}}$} & 94.27 & 2.27  & 40.10  & 3.33  & 0.029 \\
    \bottomrule
    \end{tabular}%
  }
  \label{tab:SP-ablation-depth}%
\end{table}%